\documentclass{article} 
\usepackage{nips12submit_e,times}

\usepackage{graphicx}
\usepackage{subfigure}

\usepackage{amsmath}
\usepackage{amssymb}
\usepackage{amsthm}
\usepackage{bbm}
\usepackage{enumerate}

\newcommand{\LP}{\left(}
\newcommand{\LB}{\left[}

\newcommand{\RP}{\right)}
\newcommand{\RB}{\right]}

\newcommand{\degree}{\ensuremath{^\circ}}

\newcommand{\xbf}{\ensuremath{\mathbf{x}}}

\newcommand{\PS}{P_S}
\newcommand{\PT}{P_T}

\newcommand{\DS}{D_S}
\newcommand{\DT}{D_T}

\newcommand{\RPS}{R_{\PS}}
\newcommand{\RPT}{R_{\PT}}

\newcommand{\ST}{{\langle S, T \rangle}}
\newcommand{\PST}{{P_\ST}}

\newcommand{\BST}{B_\ST}
\newcommand{\BPST}{B_\PST}

\newcommand{\RDS}{R_{\DS}}
\newcommand{\RDT}{R_{\DT}}

\newcommand{\RS}{R_{S}}

\newcommand{\Hcal}{\ensuremath{\mathcal{H}}}

\newcommand{\posterior}{\rho}
\newcommand{\prior}{\pi}

\newcommand{\des}{\operatorname{dis}_\posterior}

\newcommand{\R}{\mathbb{R}}

\newtheorem{cor}{Corollary}

\newtheorem{theorem}{Theorem}

\newcommand{\eqdef}{
    \overset{{\mbox{\rm\tiny def}}}{=}
}

\newcommand{\prob}[1]{
    \underset{#1}{\mathrm{Pr}}\
}

\newcommand{\esp}[1]{
    \underset{#1}{\mathrm{\bf E}}\
}

\newcommand{\argmindevant}[1]{
    {\mathrm{argmin}}_{#1}\
}

\newcommand{\espdevant}[1]{
    \mathrm{{\bf E}}_{#1}\,
}

\newcommand{\xb}{\mathbf{x}}

\newcommand{\KL}{{\rm KL}}
\newcommand{\kl}{{\rm kl}}

\newcommand{\loss}{\mathcal{L}}

\newcommand{\vb}{\mathbf{v}}
\newcommand{\wb}{\mathbf{w}}
\newcommand{\sgn}{\mathrm{sgn}}

\newcommand{\Phidis}{\Phi_{\rm dis}}

\title{PAC-Bayesian Learning and Domain Adaptation}

 \author{
Pascal Germain\\
D{\'e}partement d'informatique et de g\'enie logiciel\!\!\!\\ 
 Universit{\'e} Laval, Qu{\'e}bec, Canada\\
 \texttt{\small pascal.germain@ift.ulaval.ca}
\And Amaury Habrard\\
 Laboratoire Hubert Curien UMR CNRS 5516,\\
Univ. Jean Monnet, 42000 St-Etienne, France\\ 
\texttt{\small amaury.habrard@univ-st-etienne.fr}
\And Fran{\c c}ois Laviolette\\
D{\'e}partement d'informatique et de g\'enie logiciel\\ 
Universit{\'e} Laval, Qu{\'e}bec, Canada\\
 \texttt{\small francois.laviolette@ift.ulaval.ca}
\And Emilie Morvant\\
 Aix-Marseille Univ., LIF-QARMA, CNRS,\\
 UMR 7279, 13013, Marseille, France\\
 \texttt{\small emilie.morvant@lif.univ-mrs.fr}
}

\makeatletter
\def\captionof#1#2{{\def\@captype{#1}#2}}
\makeatother

\nipsfinalcopy 

\begin{document}

\maketitle

\begin{abstract}
In machine learning, {\it Domain Adaptation} (DA) arises when the distribution generating the test (target) data differs from the one generating the learning (source) data.
It is well known that DA is an hard task even under strong assumptions~\cite{BenDavid12}, among which the {\it covariate-shift} where the source and target distributions diverge only in their marginals, {\it i.e.} they have the same labeling function. 
Another popular approach is to consider an hypothesis class that moves closer the two distributions while implying a low-error for both tasks \cite{BenDavid-MLJ2010}. 
This is a \emph{VC-dim} approach that restricts the complexity of an hypothesis class in order to get good generalization. 
Instead, we propose a \emph{PAC-Bayesian} approach that seeks for suitable weights to be given to each hypothesis in order to build a majority vote.
We prove a new DA bound in the PAC-Bayesian context.
%
This  leads us to design the first DA-PAC-Bayesian algorithm based on the minimization of the proposed bound.
 Doing so, we seek for a $\posterior$-weighted majority vote that takes into account a trade-off between three quantities.
The first two quantities being, as usual in the PAC-Bayesian approach, (a) the complexity of the majority vote (measured by a Kullback-Leibler divergence) and (b) its empirical risk (measured by the $\posterior$-average errors on the source sample).  The third quantity is (c) the capacity of the majority vote to distinguish some structural difference between the source and target samples.
 \end{abstract}

\section*{Preliminaries}~\\[-15mm]

\paragraph{Domain Adaptation.} 
We consider DA for binary classification tasks where $X\!\subseteq\! \mathbb{R}^d$ is the input space of dimension $d$ and $Y\!\!=\!\{-1,1\}$ is the label set.
We have two different distributions over $X\!\times\! Y$ called the {\it source domain} $\PS$ and the {\it target domain} $\PT$. $\DS$ and $\DT$ are the respective marginal distributions over $X$. 
We tackle the challenging task where we have no information about the label on $\PT$. A learning algorithm is then provided with a {\it labeled source sample} $S\!=\!\{(\xbf^s_i,y^s_i)\}_{i=1}^{m}$ drawn {\it i.i.d.} from $\PS$, and an {\it unlabeled target sample} $T\!=\!\{\xbf^t_j\}_{j=1}^{m'}$ drawn {\it i.i.d.} from $\DT$.
Let  $h\!:\!X\!\rightarrow\! Y$ be an hypothesis function.
The {\it expected source error} of $h$ over $\PS$ is the probability that $h$ commits an error, 
$
\RPS(h) \!=\! 
\espdevant{(\xbf^s,y^s) \sim \PS} I\big( h(\xb^s) \neq y^s \big),
$
where $I(a)=1$ if predicate $a$ is true and $0$ otherwise.
The {\it expected target error} $\RPT$ over $\PT$ is defined in a similar way. $\RS$  is the {\it empirical source error}.

The DA objective is then to find a low error target hypothesis, even if no label information is available about the target domain. Clearly this task can be infeasible in general. However, under the assumption that there exists hypothesis in the hypothesis class $\Hcal$ that  do perform well on both the source and the target domain, Ben David et al.~\cite{BenDavid-MLJ2010} provide the following guarantee,
\begin{equation}
\label{eq:da}
\forall h\in\mathcal{H},\ \RPT(h)\ \leq\ \RPS(h)+\frac{1}{2}d_{{\cal H}\!\Delta\!{\cal H}}(\DS,\DT)+\nu,
\end{equation}
where $\nu\!\eqdef\!\argmindevant{h\in \mathcal{H}}(\RPS(h)+\RPT(h))$  is the error of the best joint hypothesis, and  $d_{{\cal H}\!\Delta\!{\cal H}}(D_S,D_T)$, called the ${\cal H}\!\Delta\!{\cal H}$-distance between the  domain marginal distributions, quantifies how hypothesis from $\Hcal$ can ``detect'' differences between those two distributions. According to Equation~\eqref{eq:da}, the lower this detection capability is for some given $\Hcal$, the better are the generalization guarantees. Hence, as pointed out in~\cite{BenDavid-MLJ2010},  Equation~\eqref{eq:da} together with the usual VC-bound theory,  express a multiple trade-off between the accuracy of some particular hypothesis $h$, the complexity of the hypothesis class $\Hcal$, and the ``incapacity" of hypothesis of $\Hcal$ to detect difference between the source and the target domain. 

\paragraph{PAC-Bayesian Learning of Linear Classifier.} 
The PAC-Bayesian theory, first introduced by McAllester~\cite{Mcallester99a}, traditionally considers majority votes over a set $\Hcal$ of  binary hypothesis.
Given a prior distribution~$\prior$ over $\Hcal$ and a training set $S$, the learning process consists in finding the posterior distribution $\posterior$ over $\Hcal$ leading to a good generalization.
Indeed, the essence of this theory is to bound the risk of the stochastic Gibbs classifier $G_\posterior$ associated with $\posterior$. In order to predict the label of an example~$\xbf$, the Gibbs classifier first draws a hypothesis $h$ from $\Hcal$ according to $\posterior$, then returns $h(\xbf)$ as the predicted label. Note that the error of the Gibbs classifier corresponds to the expectation of the errors over~$\posterior$: 
$R_{P_S}(G_\posterior) = \espdevant{h\sim\posterior} R_{P_S}(h)$.
The classical PAC-Bayesian theorem bounds the expectation of error $R_{P_S}(G_\posterior)$ in term of two major quantities: The empirical error $R_{S}(G_\posterior) = \espdevant{h\sim\posterior} R_{S}(h)$ on a sample~$S$ and the Kullback-Leibler divergence $\KL(\posterior\,\|\,\prior) \eqdef \espdevant{h\sim \posterior} \frac{\posterior(h)}{\prior(h)}$.
\begin{theorem}[as presented in \cite{germain2009pac}]
\label{thm:pacbayes0}
For any domain $P_S \subseteq X \times Y$, for any set $\Hcal$ of hypothesis, for any prior distribution $\prior$ over $\Hcal$, and any $\delta\in(0,1]$, we have,
\begin{equation*}
\prob{S\sim (P_S)^m}
\LP \forall \posterior \mbox{ on } \Hcal \ : \
\kl\Big(R_{S} (G_{\posterior}) \,\big\| \, R_{P_S} (G_{\posterior})\Big) \le
    \dfrac{1}{m}\LB \KL(\posterior\,\|\,\prior)+
\ln\dfrac{\xi(m)}{\delta}\RB
\RP
\ge 1 -\delta\,,
\end{equation*}
where $\kl(q\,\|\,p) \, \eqdef \, q \ln\frac{q}{p} + (1-q)\ln\frac{1-q}{1-p}$\,,
\ \ and \ \ 
$\xi(m) \, \eqdef \, \sum_{k=0}^m \binom{m}{k} \LP \frac{k}{m} \RP^{k} \LP 1 - \frac{k}{m} \RP^{m-k}\,.$
\end{theorem}

Now, let  $\Hcal$ be a set of  linear classifiers $h_\vb(\xb) \eqdef \sgn\LP\vb\cdot\xb\RP\,$ such that $\vb\in\mathbb{R}^d$ is a weight vector. 
By restricting the prior and the posterior to be Gaussian distributions,  Langford an Shawe-Taylor~\cite{Langford02} have specialized the PAC-Bayesian theory in order to bound the expected risk of any linear classifier $h_\wb\in\Hcal$ identified by a weight vector $\wb$.
More precisely, for a prior $\prior_{\mathbf{0}}$ and a posterior $\posterior_\wb$ defined as spherical Gaussians with identity covariance matrix
respectively centered on vectors $\mathbf{0}$ and $\wb$, 
{\it i.e.}
\begin{equation} \label{eq:gaussiandist}
 \mbox{\normalsize for any } h_\vb\in\Hcal\,,\quad
\prior_\mathbf{0}(h_\vb) \ \eqdef\ \LP\tfrac{1}{\sqrt{2\pi}}\RP^d
e^{-\frac{1}{2}\|\vb\|^2} 
\quad  \mbox{\normalsize and} \quad 
\posterior_\wb(h_\vb) \ \eqdef\ \LP\tfrac{1}{\sqrt{2\pi}}\RP^d
e^{-\frac{1}{2}\|\vb-\wb\|^2}\,,
 \end{equation}
we obtain that the expected risk of the Gibbs classifier $G_{\posterior_\wb}$ on a domain $P_S$ is given by,
\begin{equation*} 
R_{P_S}(G_{\posterior_\wb})  
\ = \ \esp{(\xb,y)\sim P_S}  \esp{h_\vb\sim\posterior_\wb} I(h_\vb \neq y)
\ = \ \esp{(\xb,y)\sim P_S} \Phi\left( y \,\tfrac{\wb \cdot \xb}{\|\xb\|} \right)\,,
\end{equation*}
where $\Phi(a) \ \eqdef \ \frac{1}{2} [1-\mathrm{Erf}( \mbox{\footnotesize$\frac{a}{\sqrt{2}}$} ) ]$.
Moreover, the KL-divergence between the posterior and the prior distributions becomes simply $\KL(\posterior_\wb \,\|\, \prior_\mathbf{0}) = \frac{1}{2}\| \wb \|^2$.
In this context, Theorem~\ref{thm:pacbayes0} becomes,
\begin{cor}\label{thm:pbgd1}
For any domain $P_S\subseteq \R^d \times Y$ and any $\delta\in(0,1]$, we have,
\begin{equation*}
\prob{S\sim (P_S)^m}
\LP \forall\, \wb \in \R^d \ : \
\kl\Big(R_{S} (G_{\posterior_\wb}) \,\big\| \, R_{P_S} (G_{\posterior_\wb})\Big) \le
    \dfrac{1}{m}\LB \tfrac{1}{2}\|\wb\|^2+
\ln\dfrac{\xi(m)}{\delta}\RB
\RP
\ge 1 -\delta\,.
\end{equation*}
\end{cor}
Based on this specialization of the PAC-Bayesian theory to linear classifiers, Germain et al.~\cite{germain2009pac} suggested to minimize the bound on $R_{P_S}(G_{\posterior_\wb})$ given by Corollary~\ref{thm:pbgd1}. 
The resulting learning algorithm, called PBGD, performs a gradient descent in order to find an optimal weight vector $\wb$.  Doing so, PBGD realizes a trade-off between the empirical accuracy (expressed by $R_{S} (G_{\posterior_\wb})$) and the complexity (expressed by $\|\wb\|^2$) of the learned linear classifier.


\section*{PAC-Bayesian Learning of Adapted Linear Classifier}

\paragraph{DA Bound for the Gibbs Classifier.}

The originality of our contribution is to combine PAC-Bayesian and DA frameworks. 
We define the notion of \emph{domain disagreement} $\des(\DS,\DT)$ to measure the structural difference between domain marginals in terms of posterior distribution $\posterior \sim \Hcal$,
\begin{equation*}
 \des(\DS,\DT) \ \eqdef \    \esp{h_1,h_2\sim \posterior^2 } \LB\, \RDT(h_1,h_2)  - \RDS(h_1, h_2) \,\RB\,,
\end{equation*}
where $R_{D'}(h_1, h_2) \eqdef \espdevant{\xbf\sim D'} I(h_1(\xbf)\!\ne\! h_2(\xbf))$.
Unlike the distance $d_{{\cal H}\!\Delta\!{\cal H}}$ suggested by~\cite{BenDavid-MLJ2010}, our ``distance'' measure $\des$ takes into account a $\rho$-average over all pairs of hypothesis in $\Hcal$ instead of focusing on a single particular pair of hypothesis. However, it nevertheless allows us to derive the following bound which proposes a similar trade-off as in Equation~\eqref{eq:da}, but relates the source and target errors of the Gibbs classifier. \ 
For all probability distribution $\posterior$ on $\Hcal$, we have,
\begin{equation}
\label{eq:dabound}
 \RPT(G_\posterior) \ \leq \ \RPS(G_\posterior) +  \des(\DS,\DT) + \lambda_\rho\,,
\end{equation}
where 
$\lambda_\rho\!=\! \RPT(h^\star)+\RPS(h^\star)$, with $h^\star\!=\!\argmindevant{h\in \Hcal}\{ \espdevant{h'\sim \posterior } (\RDT(h,h')\!  -\!     \RDS(h,h'))\}$, measures the joint error of the hypothesis which minimizes the domain disagreement.
Hence, similarly to Equation~\eqref{eq:da}, we provide  evidences that a good DA is possible if $\des(\DS,\DT)$ and $\lambda_\posterior$ are low. Under this assumption, we propose to design the first DA-PAC-Bayesian  algorithm inspired from the PAC-Bayesian learning of linear classifiers \cite{germain2009pac}.
We focus on 
the two first terms of Inequality~\eqref{eq:dabound}, and we refer to this quantity as the \emph{expected adaptation loss},  
\begin{equation*}
\BPST(G_\posterior) \ \eqdef \ \RPS(G_\posterior) +  \des(\DS,\DT)\,,
\end{equation*}
where $\PST$ denotes the joint distribution over $P_S \times D_T$. The independence of each draw from $P_S$ and $D_T$ allows us to rewrite $B_\PST$ as the expectation of the \emph{domain adaptation loss} $\loss_{DA}$,
\begin{eqnarray}
 B_\PST(G_\posterior) 
&=& 
\esp{h_1,h_2\sim \posterior^2}  \esp{(\xb^s,y^s,\xb^t)\sim \PST}
\loss_{DA} (h_1, h_2, \xb^s,y^s,\xb^t )\,,
\label{eq:daloss}\\[-1mm]
\loss_{DA} (h_1, h_2, \xb^s,y^s,\xb^t ) & \eqdef & I(h_1(\xb^s)\neq y^s) +   I(h_1(\xb^t) \neq h_2(\xb^t)) - I(h_1(\xb^s) \neq h_2(\xb^s))\,. 
\nonumber
\end{eqnarray}
Given  $\ST = \{(\xb^s_i, y^s_i, \xb^t_i)\}_{i=1}^m$, a sample of $m$ source-target pairs drawn \textit{i.i.d.} from $\PST$, the \emph{empirical adaptation loss} of $G_\posterior$ is $B_\ST(G_\posterior) = \espdevant{h_1,h_2\sim \posterior^2}  
\sum_{i=1}^m
\loss_{DA} (h_1, h_2, \xb^s_i,y^s_i,\xb^t_i )$.

\paragraph{PAC-Bayesian Bounds for Domain Adaptation.} We restrict ourselves to the case exhibited by Equation~\eqref{eq:gaussiandist} where $\Hcal$ is a set of linear classifiers, and posterior and prior distributions are Gaussians.
First, we compute the expected adaptation loss $ B_\PST\!(G_{\posterior_\wb})$ of the Gibbs classifier $G_{\posterior_\wb}$ (remember that the posterior distribution is centered on the linear 
$h_\wb$).
With  $\Phidis(a) \! \eqdef \! 2\Phi(a)\Phi(-a)$, we obtain,
\begin{eqnarray*}
 B_\PST(G_{\posterior_\wb}) 
&=&
\esp{(\xb^s,y^s,\xb^t)\sim\PST}
\left[
 \Phi\left( y^s \,\tfrac{\wb \cdot \xb^s}{\|\xb^s\|} \right)
 + \Phidis\left( \tfrac{\wb \cdot \xb^s}{\|\xb^s\|} \right)
 - \Phidis\left( \tfrac{\wb \cdot \xb^t}{\|\xb^t\|} \right)
 \right].
\end{eqnarray*}
Now, we derive a new PAC-Bayesian theorem to bound the expected adaptation loss of linear classifiers.  Theorem~\ref{thm:pbda_bound1} is obtained by two key results.  First, we use the specialization of the PAC-Bayesian theory to linear classifiers introduced by Corollary~\ref{thm:pbgd1}. Second, we need the methodology developed by \cite[Theorem~5]{Lacasse07} to bound a loss relying on a pair of hypothesis $h_1, h_2 \sim \posterior^2$ (like our domain adaptation loss of Equation~\eqref{eq:daloss}). We then obtain $\KL(\posterior^2_\wb\,\|\,\prior^2_\mathbf{0}) = 2\, \KL(\posterior_\wb\,\|\,\prior_\mathbf{0}) = \|\wb\|^2$.
\begin{theorem} \label{thm:pbda_bound1} 
For any domain $\PST \subseteq \R^d \times Y\times \R^d$ and any $\delta\in(0,1]$, we have,
\begin{equation*}
\prob{\ST\sim (\PST)^m}\!\!\LP\!
   \forall\, \wb \in \R^d :
\kl\Big(\BST^* \,\big\| \, \BPST^*\Big) 
\le
    \frac{1}{m}\,\LB \|\wb\|^2 +
\ln\frac{\xi(m)}{\delta}\RB
\RP \ge 1 -\delta\,,
\end{equation*}
where $\BST^* \eqdef \frac{1}{2}\BST(G_{\posterior_\wb})+\frac{1}{4}$ \ and \ $\BPST^* \eqdef \frac{1}{2}\BPST(G_{\posterior_\wb})+\frac{1}{4}$ ensure that the values provided to the $\kl(\cdot\|\cdot)$ function are in interval $[0,1]$.
\end{theorem}

%
%
%

\paragraph{Designing the Algorithm.} 
The algorithm DA-PBGD, described here, minimizes the upper bound given by 
Theorem~\ref{thm:pbda_bound1} by gradient descent. The corresponding objective function is,
\begin{equation*}\label{eq:B}
{\cal B}(\ST,\wb,\delta)\ \eqdef\ \sup\biggl\{\epsilon\,:\,
\kl(\BST^*\,\|\,\epsilon)
\ \le
    \frac{1}{m}\,\LB\|\wb\|^2 +
\ln\frac{\xi(m)}{\delta}\RB\biggr\}\, ,
\end{equation*}
for a fixed value of $\delta$. Consequently, our problem is to find weight vector
$\wb^*$ that minimizes ${\cal B}$ subject to the constraints ${\cal B} > \BST^*$ and 
$\kl(\BST^*\,\|\,{\cal B}) = \frac{1}{m}\big[ \|\wb\|^2 
  +
\ln\frac{\xi(m)}{\delta}\big]$.
The gradient is obtained by computing the partial derivative of both sides of the latter equation
with respect to $w_j$ (the $j^{th}$ component of $\wb$). After solving for
$\partial {\cal B} / \partial w_j$, we find that the gradient is,
\begin{equation*}
\tfrac{{\cal B}(1-{\cal B})}{2m({\cal B}-\BST^*)}
\Biggl[ \! 4\wb+
\ln \!\LP \tfrac{{\cal B}(1-\BST^*)}{\BST^*(1-{\cal B})} \RP \!\!
\sum_{i=1}^m\!\bigg[\! 
\Phi'\!\LP\tfrac{y^s_i\wb\cdot\xb^s_i}{\|\xb^s_i\|} \RP\! \tfrac{y^s_i\xb^s_i}{\|\xb^s_i\|} 
+ 
\Phi'_{\rm dis}\!\LP\tfrac{\wb\cdot\xb^t_i}{\|\xb^t_i\|} \RP\! \tfrac{\xb^t_i}{\|\xb^t_i\|} -
\Phi'_{\rm dis}\!\LP\tfrac{\wb\cdot\xb^s_i}{\|\xb^s_i\|} \RP\! \tfrac{\xb^s_i}{\|\xb^s_i\|} 
\bigg]\Biggr],
\end{equation*}
where $\Phi'(a)$ and $\Phidis'(a)$ denote respectively the derivatives of $\Phi$ and $\Phidis$  evaluated at $a$.
The kernel trick applied to DA-PBGD allows us to work with
dual weight vector $\pmb{\alpha}\in \R^d$ that is a linear classifier in an augmented space.  Given a kernel $k:\R^d\!\times\!\R^d\rightarrow\R$, we have $h_\wb(\xb) = \sum_{i=1}^m \alpha_i k(\xb_i, \xb)$.

\begin{figure}
\begin{tabular}{@{}c@{}c@{}c@{}c@{}}
\hspace*{-0.6cm}
\includegraphics[width=0.29\linewidth]{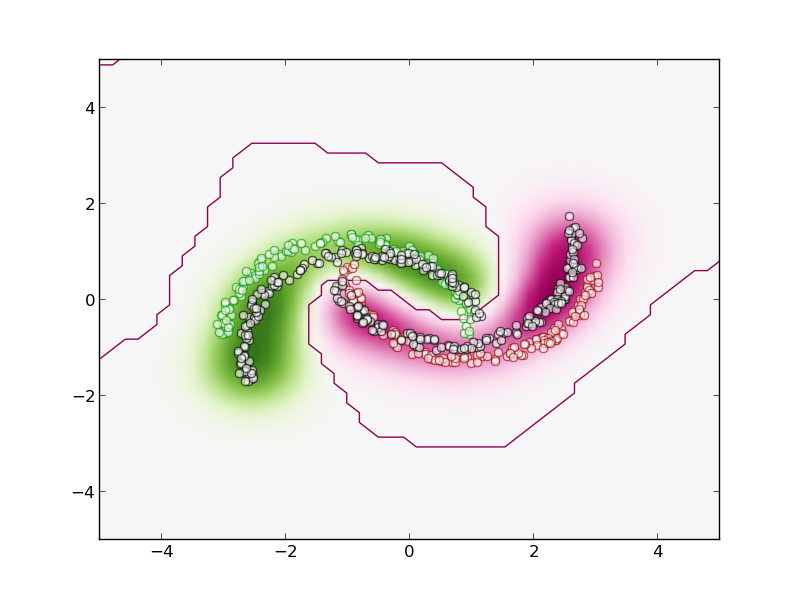}&\hspace*{-0.5cm}
\includegraphics[width=0.29\linewidth]{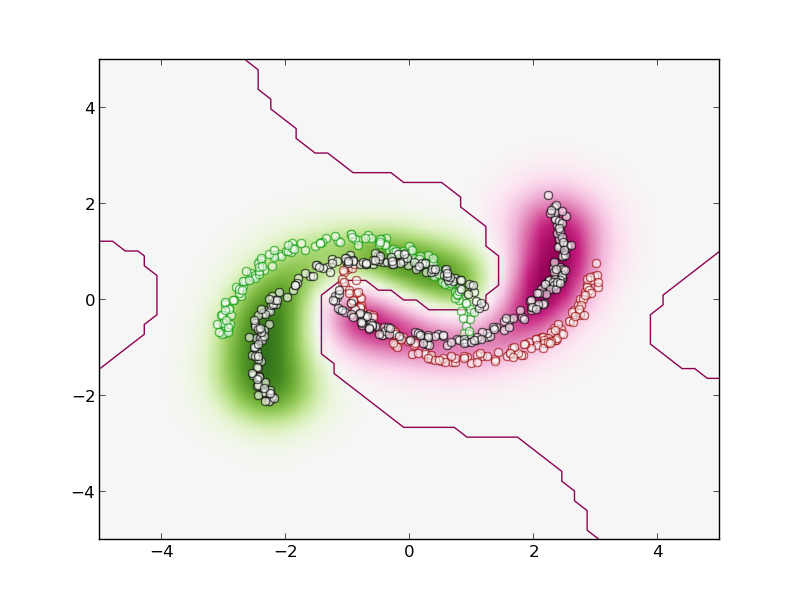}&\hspace*{-0.5cm}
\includegraphics[width=0.29\linewidth]{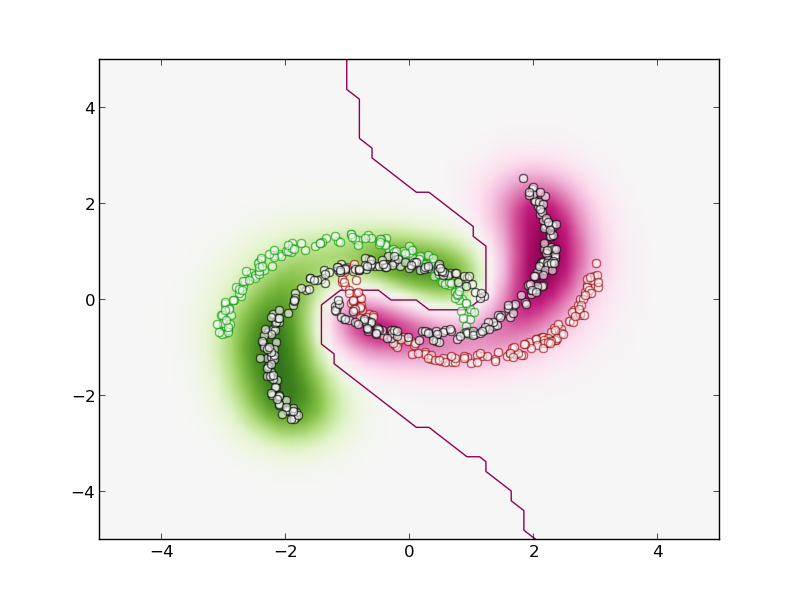}&\hspace*{-0.5cm}
\includegraphics[width=0.29\linewidth]{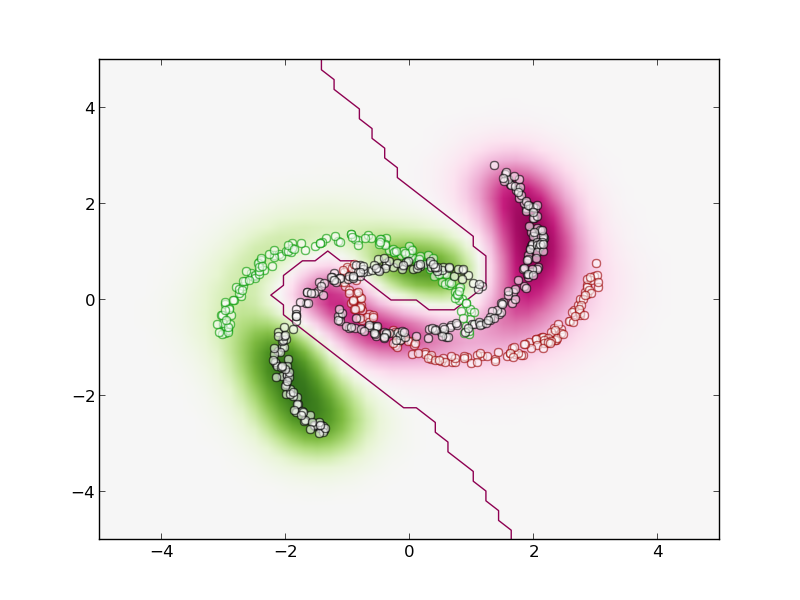}
\end{tabular}
\vspace{-4mm}
\captionof{figure}{\caption{Illustration of the decision of DA-PBGD on $4$ rotations angles: From left to right $20\degree$, $30\degree$, $40\degree$, $50\degree$. In green and pink is the source sample, in grey is the target sample. 
\label{fig:res}}}
\end{figure}

\paragraph{Experimental Results.}
Our DA-PBGD has been evaluated on a toy problem called inter-twinning moon and compared with: PBGD and  SVM with no adaptation, the semi-supervised Transductive-SVM (TSVM) \cite{Joachims99}, the iterative DA algorithms DASVM \cite{BruzzoneM10S} and the non-iterative version of DASF~\cite{DASF12} based on the bound \eqref{eq:da}. We used a Gaussian kernel for all the methods.
These preliminary results -- illustrated on Tab. \ref{tab:res} and on Fig. \ref{fig:res} -- are very promising.
Moreover on Fig. \ref{fig:tradeoff}, we clearly see the trade-off between the difficulty of the task and the minimization of the source risk in action: When the DA task is feasible DA-PBGD prefers to minimize the domain disagreement even if it implies an increase of the empirical source error, but when  this minimization becomes hard, {\it i.e.} the complexity of the task is high, it prefers to focus only on the empirical source error.

Among all the possible exciting perspectives, we notably aim to theoretically define elegant and relevant assumptions allowing one to control the $\lambda_\posterior$ term of Eq.\,\eqref{eq:dabound} to make our DA bound very~tight.

\begin{minipage}[c]{7.2cm}
\captionof{table}{\caption{Average accuracy results for $4$ rotation angles. DA-PBGD is more stable than the others and outperforms all the methods for $2$ angles. \label{tab:res}}}  
\centering
\begin{footnotesize}
 \begin{tabular}{|c||c|c|c|c|c|c|}
          \hline
          Rotation angle & $\quad20\degree\quad$\hspace*{-0.17cm} & \hspace*{-0.17cm}$\quad30\degree\quad$\hspace*{-0.17cm} &\hspace*{-0.17cm} $\quad40\degree\quad$ \hspace*{-0.17cm}& \hspace*{-0.17cm}$\quad50\degree\quad$ \hspace*{-0.17cm}\\ 
          \hline
          \hline
          PBGD           & $99.5$    & $89.8$  & $78.6$  & $60$ \\
\hline
          SVM            &  $89.6$     & $76$     &  $68.8$    &  $60$  \\ 
          \hline
          TSVM          &   $\mathbf{100}$     &  $78.9$    &  $74.6$    &  $\mathbf{70.9}$ \\ 
\hline
DASVM  & $\mathbf{100}$ & $78.4$& $71.6$&$66.6$ \\ 
 \hline
         DASF          &  $98$ & $92$ & $83$ & $70$ \\

      \hline
\hline
DA-PBGD & $97.7$  & $\mathbf{97.6}$ & $\mathbf{97.4}$ & $53.2$ \\ 
\hline    \end{tabular}
\end{footnotesize}
\end{minipage}
\hspace{8mm}
\begin{minipage}[c]{5.7cm}
\centering
 \includegraphics[height=2.4cm]{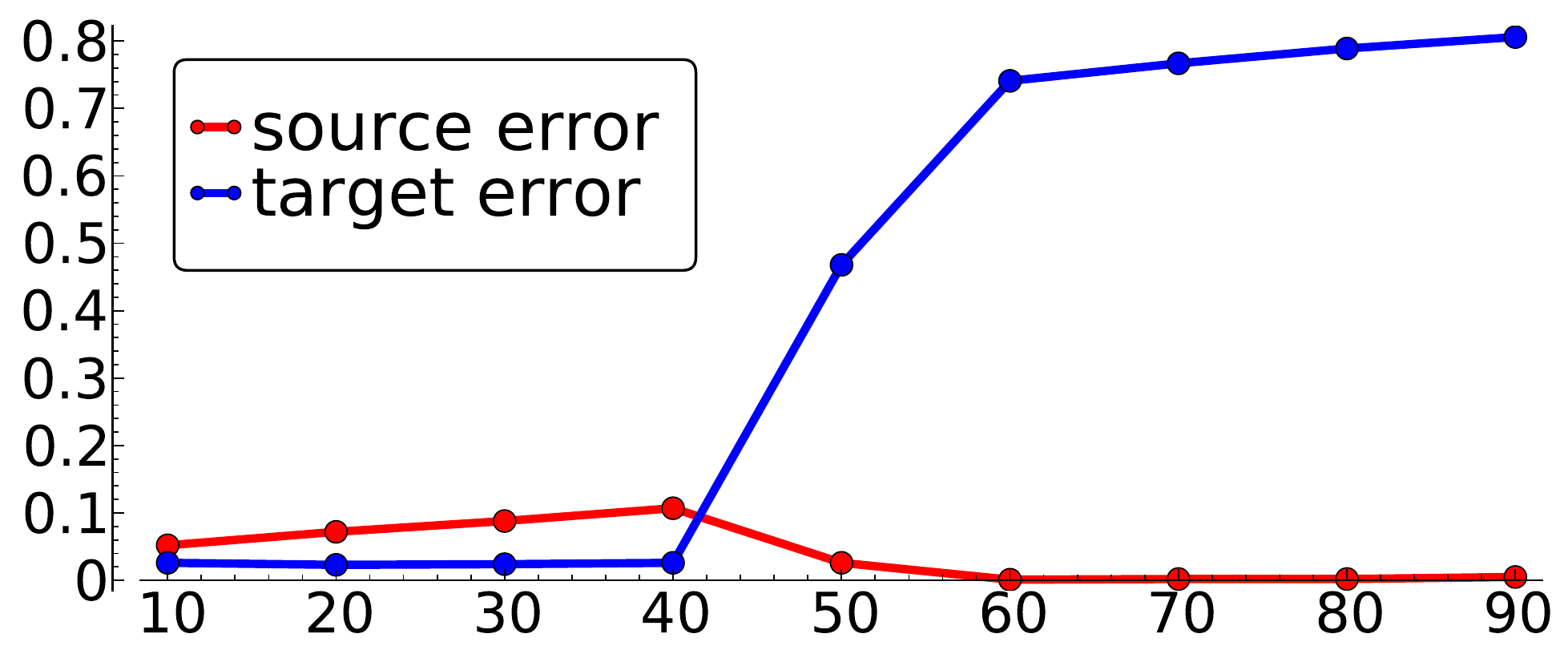}
\vspace{-6mm}
\captionof{figure}{\caption{The trade-off between target and source errors according to the difficulty of the task ({\it i.e.} the rotation angle).
\vspace{-8mm}}\label{fig:tradeoff}}
\end{minipage}

\begin{footnotesize}
\paragraph{Acknowledgments}
This work was supported in part by the french project VideoSense {\scriptsize ANR-09-CORD-026}, in part by the IST Programme of the European Community, under the PASCAL2 Network of Excellence {\scriptsize IST-2007-216886} and in part by NSERC discovery grant {262067}. This publication only reflects authors' views.
\end{footnotesize}

\renewcommand{\refname}{\vspace{-2em}}
\subsubsection*{References}
{\small
\bibliography{biblio}

\begin{thebibliography}{1}

\bibitem{BenDavid12}
S.~Ben-David and R.~Urner.
\newblock On the hardness of domain adaptation and the utility of unlabeled
  target samples.
\newblock In {\em Proceedings of Algorithmic Learning Theory}, pages 139--153,
  2012.

\bibitem{BenDavid-MLJ2010}
S.~Ben-David, J.~Blitzer, K.~Crammer, A.~Kulesza, F.~Pereira, and J.W. Vaughan.
\newblock A theory of learning from different domains.
\newblock {\em Machine Learning Journal}, 79(1-2):151--175, 2010.

\bibitem{Mcallester99a}
David~A. McAllester.
\newblock Some {PAC}-bayesian theorems.
\newblock {\em Machine Learning}, 37:355--363, 1999.

\bibitem{germain2009pac}
P.~Germain, A.~Lacasse, F.~Laviolette, and M.~Marchand.
\newblock {PAC-Bayesian Learning of Linear Classifiers}.
\newblock In {\em Proceedings of ICML}, 2009.

\bibitem{Langford02}
J.~Langford and J.~Shawe-Taylor.
\newblock {PAC}-bayes \& margins.
\newblock In {\em Advances in Neural Information Processing Systems 15}, pages
  439--446. MIT Press, 2002.

\bibitem{Lacasse07}
Alexandre Lacasse, Fran\c{c}ois Laviolette, Mario Marchand, Pascal Germain, and
  Nicolas Usunier.
\newblock {PAC}-bayes bounds for the risk of the majority vote and the variance
  of the {G}ibbs classifier.
\newblock In {\em NIPS}, 2007.

\bibitem{Joachims99}
T.~Joachims.
\newblock Transductive inference for text classification using support vector
  machines.
\newblock In {\em ICML}, 1999.

\bibitem{BruzzoneM10S}
L.~Bruzzone and M.~Marconcini.
\newblock Domain adaptation problems: A {DASVM} classification technique and a
  circular validation strategy.
\newblock {\em IEEE Trans. Pattern Anal. Mach. Intell.}, 32(5), 2010.

\bibitem{DASF12}
E.~Morvant, A.~Habrard, and S.~Ayache.
\newblock {Parsimonious Unsupervised and Semi-Supervised Domain Adaptation with
  Good Similarity Functions}.
\newblock {\em {Knowledge and Information Systems}}, 33(2):309--349, 2012.

\end{thebibliography}
\bibliographystyle{unsrt}
}

\end{document}